# A Weakly-Supervised Depth Estimation Network Using Attention Mechanism

Fang Gao[1], Jiabao Wang[1], Jun Yu[2]*, Yaoxiong Wang[3], Feng Shuang[1]

[1]School of Electrical Engineering, Guangxi University, Nanning 530004, China
[2]Department of Automation, University of Science and Technology of China, Hefei 230027, China
[3]Institute of Intelligent Machines, Chinese Academy of Sciences, Hefei 230031, China
fgao@gxu.edu.cn, 601533944@qq.com, harryjun@ustc.edu.cn, yxwang@iim.ac.cn, fshuang@gxu.edu.cn

## Abstract

Monocular depth estimation (MDE) is a fundamental task in many applications such as scene understanding and reconstruction. However, most of the existing methods rely on accurately labeled datasets. A weakly-supervised framework based on attention nested U-net (ANU) named as ANUW is introduced in this paper for cases with wrong labels. The ANUW is trained end-to-end to convert an input single RGB image into a depth image. It consists of a dense residual network structure, an adaptive weight channel attention (AWCA) module, a patch second non-local (PSNL) module and a soft label generation method. The dense residual network is the main body of the network to encode and decode the input. The AWCA module can adaptively adjust the channel weights to extract important features. The PSNL module implements the spatial attention mechanism through a second-order non-local method. The proposed soft label generation method uses the prior knowledge of the dataset to produce soft labels to replace false ones. The proposed ANUW is trained on a defective monocular depth dataset and the trained model is tested on three public datasets, and the results demonstrate the superiority of ANUW in comparison with the state-of-the-art MDE methods.

## 1 Introduction

Depth information estimation from 2D images is a key step in scene reconstruction and understanding tasks, such as 3D object recognition, segmentation and detection. In this paper, monocular depth estimation (MDE) from a single RGB image is studied on a defective monocular depth dataset and other two normal datasets.

Compared with the depth estimation from stereo images or video sequences [Ha et al., 2016; Karsch et al., 2014], the progress of MDE is relatively slow due to its highly underdetermined nature, that is, a single RGB image may be acquired from infinitely different 3D scenes. In order to overcome this inherent uncertainty, depth-related features, such as perspective and texture information, object size, object position and occlusion, have to be utilized in the methodological development [Hoiem et al., 2007; Ladicky et al., 2014].

---

* Corresponding Author

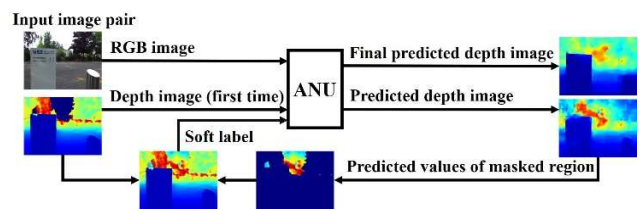

Figure 1. The proposed ANUW framework. The RGB and depth image are fed into the ANU network for training. The predicted depth image is used to iteratively update the soft label, which is obtained as the ground-truth depth image in the new iteration by replacing depth values of masked region with predicted ones. Training with the iteratively updated soft label can improve the prediction accuracy.

Some recent works based on deep convolution neural network (DCNN) [Eigen and Fergus., 2015; Kim et al., 2016] take the task of MDE as a supervised learning problem and show significant improvement, which verifies that the features extracted by neural network is much better than the ones extracted by hand. Other methods [Geiger et al., 2013; Alhashim et al., 2018] solve the MDE problem by training the deep fully convolution neural network, which is originally designed as the feature extractor for image classification, to predict the continuous depth map. In these networks, repeated spatial pooling operations will quickly reduce the spatial resolution of feature maps, which is undesirable for depth estimation. Although the resolution can be improved by fusing high-resolution low-level features through multi-layer deconvolution network, multi-scale network or jump connections, such processing makes the network structure and training process more complicated together with additional computation and memory cost. The regression network has no such problems, but it has the disadvantages of slow convergence speed and poor visualization.

Unsupervised learning methods can be employed when video sequences or multi-view information are available [Kuznietsovet al., 2017; Godard et al., 2017]. In these methods, 2D frames are firstly projected back to the 3D space by utilizing time or space information, and the projected frames are aligned in the 3D space according to the estimated depth and the camera's relative pose. Without ground-truth depth supervision, unsupervised methods can estimate the depth by minimizing a photometric loss after the projection of the estimated 3D information again to 2D. Recent approaches adopted weakly-supervised training [Garg et al., 2016], where the network is optimized based on the ordinal (i.e. farther or closer) relationship between pixel pairs. However, the accuracy of these methods is limited, since the pixel-pair based labels are very sensitive to noise.

We improved the depth estimation quality with a novel weakly-supervised framework named as ANUW. An example result of our method is shown in Fig. 1.

The contributions of this paper are highlighted below:
- A network named as ANU is proposed based on the U-net [Ronneberger et al., 2015] structure. ANU uses long and short jump connections to retain the detailed underlying information, and adds two types of self-attention modules (i.e. AWCA and PSNL) to selectively emphasize channel and spatial features to extract better contextual dependencies.
- A soft label generation method is proposed to show that training with the generated soft labels method performs better than training with error ones. Updating labels of masked region for the next iteration with the current output can make use of the robustness of the model to reduce the proportion of wrong labels.
- A training loss function is proposed to further improve the performance by minimizing depth value errors, gradient errors and structural similarity (SSIM)errors.

## 2 Related Work

### 2.1 Feature Extraction Network

Depth Estimation is essential to understand the three-dimensional structure of the scene from two-dimensional images, and feature extraction is one key step in depth estimation. Early work developed geometry-based methods [Flynn et al., 2016] to predict depth from stereo images. In view of the success of DCNN in image understanding, many depth estimation networks have been proposed in recent years. Based on the multi-level contextual and structural information captured by powerful DCNN (such as very deep convolutional networks (VGG) [Simonyan et al., 2014] and residual network (ResNet) [He et al., 2016]), the accuracy of depth estimation has been raised to a new level. [Laina et al., 2016] proposed a fully convolutional residual network (FCRN), which utilized small convolutions instead of large ones in ResNet-50 to achieve up-sampling and get better results. For MDE in video, the DeepV2D method proposed by [Teed et al., 2018] combines two classical methods in an end-to-end architecture, where the depth estimation module takes the camera motion as the input and returns the initial depth map while the camera motion module obtains the predicted depth and outputs the accurate camera motion trajectory. With the development of transfer learning, some neural networks with excellent performance in the field of semantic segmentation are used as pre-training models for depth estimation. Due to its excellent feature extraction ability, U-net is a potential candidate as the encoder part of different neural networks.

### 2.2 Attention Mechanism

In general, attention mechanism can be viewed as a tool to redistribute available information and focus on the salient components of an image, and it has already played an important role in the current computer vision society, such as video classification, super-resolution and scene segmentation.

[Vaswani et al., 2017] is the first work to replace recurrent neural network with the scaled dot-product attention module in the sequential model to obtain the weights. With the development of attention mechanism, spatial attention and channel attention modules are proposed to extract deeper features. In specific, [Xia et al., 2019] presented a novel attention module in spatial domain incorporating non-local operations with second-order statistics in convolutional neural network (CNN) to extract contextual dependencies. [Hu and Sun., 2018] proposed a squeeze-and-excitation block in channel domain to model channel-wise feature correlations for image classification. [Fu et al., 2019] proposed a dual attention network that includes spatial and channel attention modules, and the features outputted by the attention network are fused and then fed into the semantic segmentation network. [Li et al., 2020] proposed the AWCA and PSNL modules to improve the performance of hyperspectral reconstruction.

### 2.3 Weakly-Supervised Learning (WSL)

With the development of depth estimation, unsupervised and weakly-supervised learning is introduced into the depth estimation network. The recent monocular video depth estimation method Monodepth2 [Godard et al., 2019] uses U-net and VGG as the depth and camera pose estimation networks, respectively. The estimated camera pose is employed to supervise the training of the depth estimation network.

Pseudo labeling based methods [Lee et al., 2013] improve the performance of WSL by generating high-quality hard labels (i.e., the arg max of the output class probability) of unlabeled data with a predefined threshold and then retraining the model with these labels. Some other works solve the depth estimation problem by using weakly labeled pixel pairs. [Chen et al., 2016] manually labeled such pixel pairs in web-collected images and constructed the Depth in the Wild dataset. Then an hourglass network is trained to generate relative depth maps based on these pixel pair labels. [Li and Noah., 2018] used the structure-from-motion method to generate the ground-truth depth information for landmarks, and further extracted the weakly labeled pixel pairs from foregrounds and landmarks separately. However, the accuracy of these weakly supervised methods is limited since the pixel-level depth estimation is very sensitive to noise.

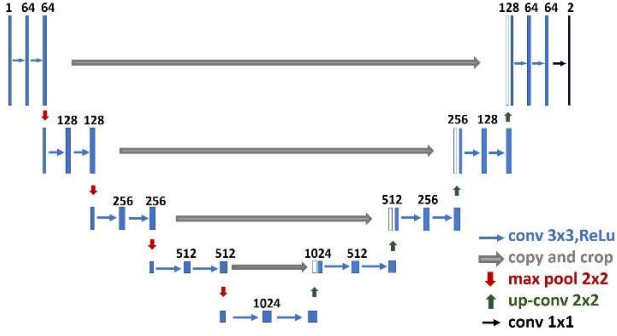

Figure 2. The structure of U-net. The number of channels is labelled for each module in the network.

## 3 Proposed Method

This section introduces our end-to-end depth estimation neural network ANUW in details, which includes its architecture, the added AWCA and PSNL modules, the novel loss function and the soft label generation method.

### 3.1 Network Architecture and Attention Modules

**ANU.** Previous depth estimation networks usually use the standard DCNN originally designed as the feature extractor for image recognition. However, the combination of max-pooling and deep convolution significantly reduces the spatial resolution of the feature map, which is undesirable for depth estimation.

max-pooling and deep convolution significantly reduces the spatial resolution of the feature map, which is undesirable for depth estimation.

As shown in Fig.2, the recently proposed U-net consisting of down-sampling, up-sampling and skipping connections can reconstruct high-resolution depth maps by fusing low-level information. However, due to the lack of inter-layer connections, the low-level features are not fully utilized.

As shown in Fig.3(a), more inter-layer connections are added to fuse multi-scale information in our ANU network, which is composed of down-sampling, up-sampling, inter-layer connections and dense residual connections. The multi-scale image features can be fused by the inter-layer skipping connections and the basic features can be preserved to the greatest extent by the dense residual connections. The features output by different modules are fused by concatenation.

**AWCA Module.** Exploiting the correlation of channel-wise features is essential to improve discriminative learning ability of the network. The AWCA module shown in Fig.4 can selectively emphasize information features by learning adaptive channel-wise feature weights.

In this module, the input $F = [f_1, f_2, \cdots, f_c, \cdots, f_C]$, with $f$ denoting a feature map with size of $H \times W$, is firstly reshaped to $R^{C \times (H \times W)}$. A convolutional layer is then used to learn the adaptive weighting matrix $Y \in R^{1 \times H \times W}$, which is reshaped to $R^{(H \times W) \times 1}$ and normalized by the softmax layer. The normalized $Y$ is multiplied with F to obtain the channel descriptor:

$$Z = [z_1, z_2, \cdots, z_C] \quad (1)$$

The above process is called adaptive weighted pooling. The channel descriptor is fed into the sequential convolutional layers followed by the ReLU and Sigmoid operations,

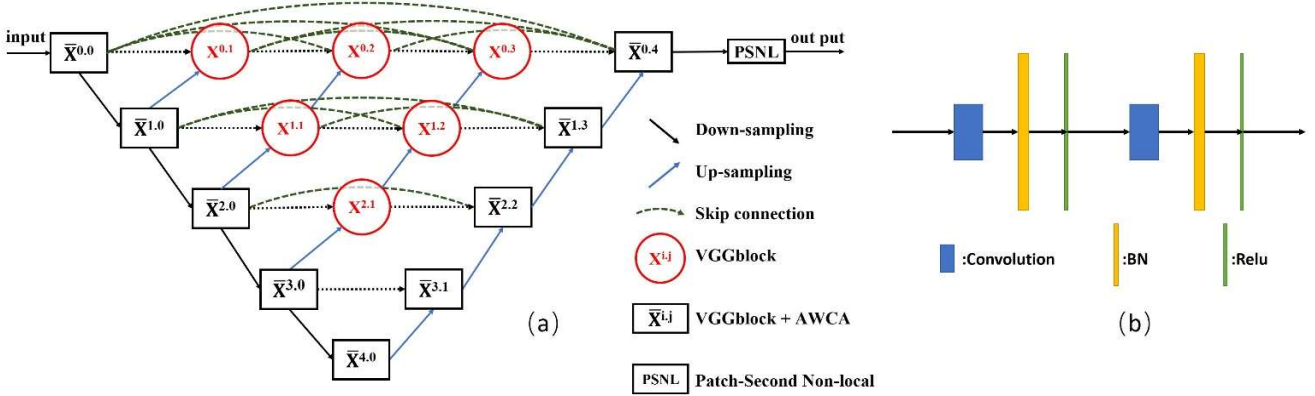

Figure 3. (a) The structure of ANU; (b) The structure of VGG block used in ANU.

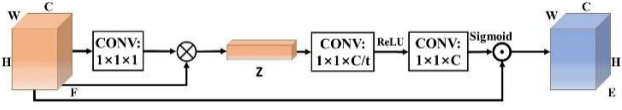

Figure 4. The AWCA module. ⊙ denotes element-wise multiplication.

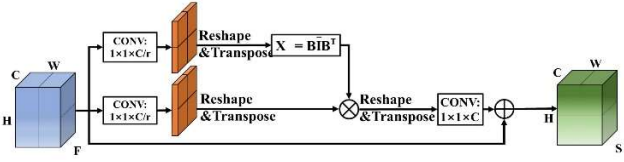

Figure 5. The PSNL module. ⊗ denotes matrix multiplication.

respectively, and the resulted channel mapping is:

$$V = \delta\left(W_2\left(\sigma\left(W_1(Z)\right)\right)\right) \quad (2)$$

where $W_1$ and $W_2$ are the weight sets of two convolutional layers, and $\delta(\cdot)$ and $\sigma(\cdot)$ represents Sigmoid and ReLU activation functions. The output channels of these two convolutional layers are $C/t$ and $C$, respectively. Finally, $V = [v_1, v_2, \cdots, v_c, \cdots, v_C]$ is used to rescale the input F to obtain the channel-wise feature map $E$:

$$E = [e_1, \cdots, e_c, \cdots] = [v_1 \cdot f_1, \cdots, v_c \cdot f_c, \cdots] \quad (3)$$

here the subscript $c$ means the $c$-th channel. The AWCA module is embedded in ANU to adjust channel-wise feature recalibration adaptively and thus boost representational learning of the network.

**PSNL Module.** It has been shown that second-order statistics is effective to enhance discriminative representation ability of CNN. The PSNL module shown in Fig.5 can model distant region relationships without bringing huge computational burden. A feature map $F \in R^{C \times H \times W}$ is firstly splitted into four sub feature maps $F_k \in R^{C \times h \times w}$ ($k=1, 2, 3, 4; h = H/2; w = W/2$) along two spatial dimensions, and each sub feature map is then processed by the PSNL module.

In the PSNL module, the feature map $F_k \in R^{C \times h \times w}$ is fed into a 1×1 convolutional layer with output channel = $C/r$ to obtain two new feature maps $B_k$ and $D_k$. Then $B_k$ is used to obtain the spatial attention map $X_k$:

$$X_k = B_k \bar{I}_k B_k^T \quad (4)$$

where $\bar{I} = \frac{1}{n}\left(I - \frac{1}{n}I\right)$ and $n = h \times w$. $X_k \in R^{n \times n}$ is fed into a softmax layer and the output is multiplied by $D_k$:

$$U_k = softmax(X_k) D_k \quad (5)$$

$U_k$ is reshaped and transposed to $R^{h \times w \times C/r}$ and fed into a 1 × 1 convolutional layer $\varphi(\cdot)$, and the output feature is added to the original feature $F_k$ by the residual connection

$$S_k = \varphi(U_k) + F_k \quad (6)$$

The resulted new feature map S contains rich spatial contextual information. As shown in Fig.3(a), the PSNL module is added as the final step in our ANU network to extract features containing spatial contextual information.

### 3.2 The Loss Function

A standard loss function for depth regression problems considers the difference between the ground-truth depth map $y$ and the predicted depth map $\hat{y}$. There are many different loss functions employed in the depth estimation neural network. Different considerations in loss functions will lead to different effects on the training speed and the overall depth estimation performance. For example, gradient loss can focus on areas with large changes and improve the prediction quality of these areas. In our method, a loss function with $L_1$ loss, gradient loss and structural similarity (SSIM) loss is proposed. This novel loss function tries to reduce the final reconstruction error by minimizing depth value errors, gradient errors and structural similarity errors. The novel loss function $L(y,\hat{y})$ is defined as the weighted sum of three terms:

$$L(y,\hat{y}) = \lambda_1 L_{depth}(y,\hat{y}) + \lambda_2 L_{grad}(y,\hat{y}) + \lambda_3 L_{SSIM}(y,\hat{y}) \quad (7)$$

The first term $L_{depth}$ is the point-wise $L_1$ loss depending on the depth values:

$$L_{depth}(y,\hat{y}) = \frac{1}{n}\sum_p^n |y_p - \hat{y}_p| \quad (8)$$

The second term $L_{grad}$ is the $L_1$ loss defined with the gradients of the depth image:

$$L_{grad}(y,\hat{y}) = \frac{1}{n}\sum_p^n |g_x(y_p,\hat{y}_p)| + |g_y(y_p,\hat{y}_p)| \quad (9)$$

where $g_x$ and $g_y$ are, respectively, the point-wise gradient differences of y and $\hat{y}$ along $x$ and $y$ axes.

The last term $L_{SSIM}$ is a commonly-used metric in image reconstruction tasks and has also been recently shown to be a useful metric for depth estimating CNN. When SSIM achieves its upper bound 1, $L_{SSIM}$ is minimized to 0:

$$L_{SSIM}(y,\hat{y}) = \frac{1 - SSIM(y,\hat{y})}{2} \quad (10)$$

In our experiment, the weights in the loss function are taken to be $\lambda_1=0.2, \lambda_2=0.3, \lambda_3=0.5$.

### 3.3 Soft Label Generation

Since there are a large number of error labels in our defective monocular depth dataset, a weakly-supervised learning framework ANUW shown in Fig.6 is employed to generate soft labels for depth estimation. This framework can make full use of the robustness of ANU network and iteratively correct the error labels. In our defective dataset, the depth values are obtained based on light reflection, so values of certain areas such as the sky can not be obtained, and pixels without depth information were incorrectly labeled with depth value 0. These wrong labels are then replaced with the corresponding generated soft depth labels in each iteration.

In the soft label generation process, a mask matrix W is used to mark the region with the labeled depth value 0 in the training set. Then the matrix W is used to filter the predicted depth map to obtain the depth values of the corresponding regions as the soft labels. These soft labels together with those of the other regions that are correctly labeled are used as the new ground-truth depth image. If the loss is reduced in the new training iteration, labels of masked region in this iteration will be updated. The updating approach can make the reconstruction error drop quickly, and the network has learned the best weights when the loss no longer decreases.

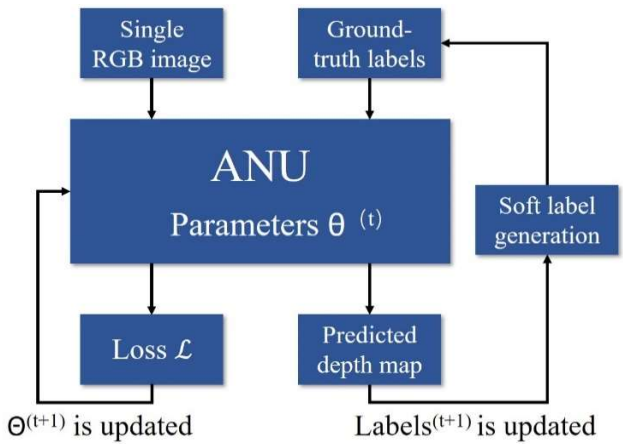

Figure 6. ANUW Framework.

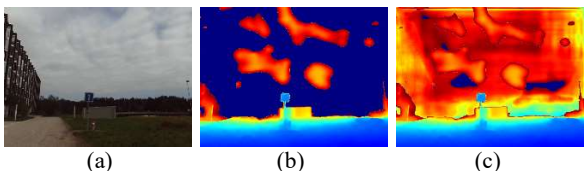

(a)          (b)          (c)

Figure 7. Monocular depth dataset. (a)The RGB image; (b)The corresponding depth image without soft labels; (c) The corresponding depth image with soft labels.

As mentioned above, our defective monocular depth dataset has many incorrect labels, especially in the sky area (see in Fig.7(b)). The weakly-supervised learning can iteratively correct the wrong pixel-level depth labels. As shown in the Fig.7(c), the wrong labels of the sky area are replaced with the generated soft labels, which are closer to the true depth values.

## 4 Experiments

### 4.1 Settings

**Monocular Depth dataset.** This paper uses the Mai depth dataset provided in the Mobile AI MDE challenge [Ignatov and Timofte., 2021] as the training dataset. The challenge aims to predict depth maps from RGB images as accurate as possible. The evaluation indexes include reconstruction accuracy and speed. The Mai depth dataset consists of 7385 RGB and depth image pairs with the size of 640×480. The RGB image has three channels while the depth image has single channel. The dataset is divided into a training set of 7208 image pairs and a verification set of 177 image pairs. KITTI [Geiger et al., 2013] dataset and DIODE [Vasiljevic et al., 2019] dataset are also used to test the trained model. The KITTI dataset contains a large number of road scenes, while the DIODE dataset scenes are mostly playgrounds and buildings.

**Evaluation index.** Three indexes, that are absolute relative error (Abs Rel) in Eq. (11), SSIM in Eq. (12) and scale invariant root mean squared error (Si-RMSE) in Eq. (13), are used in our experiments to quantify the reconstruction accuracy.

$$Abs\ Rel = \frac{1}{N} \sum \left| \frac{y - \hat{y}}{\hat{y}} \right| \quad (11)$$

$$SSIM(y,\hat{y}) = \frac{(2\mu_y \mu_{\hat{y}} + c_1)(2\sigma_{y\hat{y}} + c_2)}{(\mu_y^2 + \mu_{\hat{y}}^2 + c_1)(\sigma_y^2 + \sigma_{\hat{y}}^2 + c_2)} \quad (12)$$

$$Si - RMSE = \sqrt{\frac{\sum_{i,j=0}^{h,w} (\hat{y} - y)^2}{N} - \frac{(\sum_{i,j=0}^{h,w} (\hat{y} - y))^2}{N^2}} \quad (13)$$

where $y$ is the ground-truth depth map and $\hat{y}$ is the predicted depth map, $N$ represents the number of pixels contained in $y$, $\mu$ represents the average value, $\sigma$ represents the variance, $\sigma_{y\hat{y}}$ represents the covariance between $y$ and $\hat{y}$, $c_1$ and $c_2$ are constants.

**Implementation details.** Our ANU network has five layers, four down-sampling operations, four up-sampling operations and nine AWCA modules. In the training process, the batch number is set to 1, and the Adam optimizer is employed with the parameters $\beta_1 = 0.9$, $\beta_2 = 0.99$ and $\epsilon = 10^{-8}$. In AWCA, the scaling ratio $t$ is set to 16. In PSNL, the value of $r$ is set to 8. The initial learning rate is 0.0001, and the attenuation coefficient of the polynomial function is 1.5. The maximum epoch is set to 100. Our proposed ANU network has been implemented in Pytorch, and run on an NVIDIA 2080 Ti GPU.

### 4.2 Ablation Experiments

Ablation experiments on the Mai depth dataset are conducted to verify the effect of different modules. Table 1 lists the results. Here A is the U-net++ network; B and C add the AWCA and PSNL modules to the U-net++ network, respectively; D is ANU with both the AWCA and the PSNL modules; E is ANUW with weakly-supervised learning for soft label generation.

**AWCA.** It can be seen from Table 1 that the Si-RMSE of A is 0.612, while that of B decreases to 0.609. The main reason is that the AWCA module adaptively changes the weight of different channels and improves the learning ability of the network.

| Method | A | B | C | D | E |
|---|---|---|---|---|---|
| AWAC | × | √ | × | √ | √ |
| PSNL | × | × | √ | √ | √ |
| Soft labels | × | × | × | × | √ |
| Si-RMSE | 0.612 | 0.609 | 0.604 | 0.597 | 0.582 |

Table 1 Results of Ablation Experiments.

| Method/dataset | Mai depth | | | KITTI | | | DIODE | | |
|---|---|---|---|---|---|---|---|---|---|
| | SSIM | Abs Rel | Si-RMSE | SSIM | Abs Rel | Si-RMSE | SSIM | Abs Rel | Si-RMSE |
| **U-net** [Ronneberger et al., 2015] | 0.481 | 0.578 | 0.628 | 0.809 | 0.433 | 0.816 | 0.251 | 0.944 | 0.831 |
| **U-net++** [Zhou et al., 2018] | 0.588 | 0.773 | 0.612 | 0.846 | 0.293 | 0.699 | 0.381 | 0.732 | 0.731 |
| **FCRN** [Laina et al., 2016] | 0.613 | **0.553** | 0.605 | 0.711 | 0.178 | 0.845 | 0.399 | 0.549 | 0.682 |
| **Monodepth2** [Godard et al., 2019] | 0.415 | 0.665 | 0.595 | 0.603 | **0.156** | 0.473 | 0.305 | 0.652 | 0.693 |
| **ANUW** | **0.629** | 0.669 | **0.582** | **0.875** | 0.165 | **0.449** | **0.405** | **0.503** | **0.621** |

Table 2 Test results of four other monocular depth estimation methods and our proposed framework on three public datasets. The best results have been shown in bold.

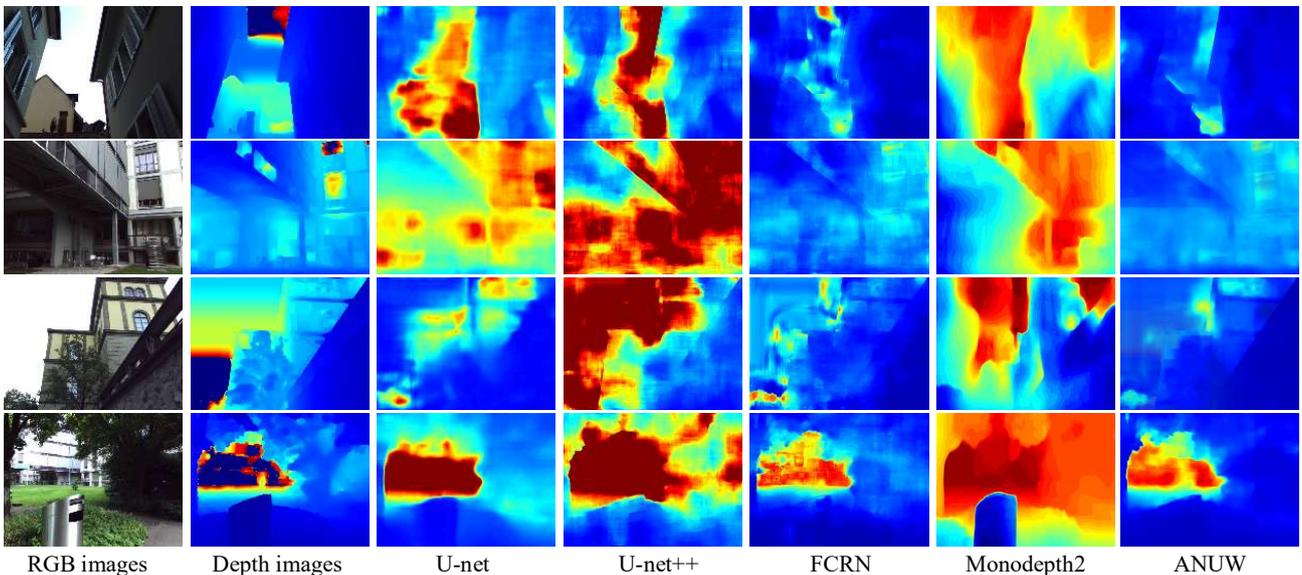

RGB images   Depth images   U-net   U-net++   FCRN   Monodepth2   ANUW

Figure 8. The predicted depth maps with different methods. The second column gives the ground-truth depth images.

**PSNL.** The Si-RMSE of A and C show that this module can make use of the spatial context information to achieve a higher reconstruction accuracy at the cost of adding a small number of parameters. The Si-RMSE can be further reduced to 0.597 when both the AWCA and PSNL modules are added.

**Soft-labels.** By adding the above two modules and updating the soft labels by weakly-supervised learning, ANUW can achieve the best performance, and the Si-RMSE is reduced to 0.582.

### 4.3 Results

In order to verify the superiority of our proposed weakly-supervised framework ANUW, we compare our method with U-net, U-net++, FCRN and Monodepth2. Table 2 lists the results on the three public datasets.

The results show that our method is better than the other four methods in the metrics of SSIM, Abs Rel and Si-RMSE. Some RGB images and their corresponding depth maps predicted with different methods are listed in Fig.8. It can be seen that the depth maps predicted by our method are closest to the truth. Therefore, our ANUW method achieves the best performance compared to the other three methods.

## 5 Conclusion

In this paper, an improved ANUW framework for MDE is proposed. The employed ANU network is a modified U-net with dense residual connections, skipping inner-layer and inter-layer connections, and channel and spatial attention modules. Our framework has four advantages: (1) In order to obtain high-resolution depth maps, multi-scale feature fusion and dense residual connections are performed in the network, which enables to capture more details and thus predict better depth maps; (2) The extracted features are utilized more efficiently with the AWCA and PSNL modules, and the pixel-level prediction accuracy is improved; (3) A novel loss function with three loss terms is proposed, and the weights can be fine-tuned to account for bias among different metrics; (4)

The soft label generation method eliminates the influence of mislabeling and helps to predict more realistic depth maps.

By comparing with some existing methods on the three public datasets, it is shown that our method can significantly improve the depth estimation accuracy. Our method can also be transferred to the field of depth image inpainting and pose estimation. As the depth values of our dataset are imbalanced (most depth values concentrate on the middle range, while the short/long range values are sparse), so the prediction accuracy can be further improved by introducing deep imbalanced regression modules [Yuzhe et al., 2021] in addition to optimizing the network structure.

## Acknowledgements

This work was supported by the Guangxi Science and Technology Base and Talent Project under Grant 2020AC19253, the Natural Science Foundation of China under Grant 61773359 and Grant 61720106009, Anhui Province Key Research and Development Program (202104a05020007) and USTC Research Funds of the Double First-Class Initiative (YD2350002001).